\documentclass[runningheads]{llncs}
\usepackage[T1]{fontenc}
\usepackage{graphicx}
\usepackage{subfig} 
\usepackage{caption}
\usepackage{booktabs}  
\usepackage{amssymb}
\usepackage{enumitem}
\usepackage{pifont}
\usepackage{tabularx}
\usepackage{booktabs}

\begin{document}
\title{Retrieval-Augmented VLMs for Multimodal Melanoma Diagnosis}

\author{Jihyun Moon\orcidID{0009-0009-4373-2600} \and
Charmgil Hong\orcidID{0000-0002-8176-252X}}

\authorrunning{Moon and Hong}

\institute{Handong Global University, Pohang, Republic of Korea\\
\email{\{jhmoon, charmgil\}@handong.ac.kr}}

\maketitle             

\begin{abstract}
Accurate and early diagnosis of malignant melanoma is critical for improving patient outcomes. While convolutional neural networks (CNNs) have shown promise in dermoscopic image analysis, they often neglect clinical metadata and require extensive preprocessing. Vision-language models (VLMs) offer a multimodal alternative but struggle to capture clinical specificity when trained on general-domain data. To address this, we propose a retrieval-augmented VLM framework that incorporates semantically similar patient cases into the diagnostic prompt. Our method enables informed predictions without fine-tuning and significantly improves classification accuracy and error correction over conventional baselines. These results demonstrate that retrieval-augmented prompting provides a robust strategy for clinical decision support.

\end{abstract}

\section{Introduction}
Malignant melanoma is the most common and deadliest form of skin cancer, with 100,640 new cases and 8,290 deaths reported in the United States in 2024~\cite{siegel2024cancer}. 
Early detection significantly improves survival, which emphasizes the need for accurate and timely diagnosis. 
Automated diagnostic tools can assist clinicians in detecting malignant lesions at an earlier stage. This can lead to improved prognosis and make timely intervention more achievable. 
While convolutional neural network (CNN)-based methods have shown promise~\cite{esteva2017dermatologist,han2018classification}, most rely solely on dermoscopic images and often require preprocessing steps such as region of interest (ROI) segmentation, which limits their utility in clinical practice.

To address these limitations, recent efforts have explored multimodal frameworks that incorporate both images and clinical metadata have gained attention for improving diagnostic accuracy and personalization. 
Vision-language models (VLMs)~\cite{liu2023visual,akrout2024evaluation} have emerged as strong candidates for such tasks, as they jointly process visual and textual data without the need for handcrafted preprocessing. 
However, off-the-shelf VLMs that are trained on general-purpose data often fail to capture domain-specific complexities~\cite{chen2024can}. 
Although fine-tuning with clinical data can mitigate this issue, it requires curated datasets and significant computational resources. 
As a result, this approach is often infeasible due to privacy constraints and institutional variability.

As an alternative, retrieval-augmented generation (RAG)~\cite{lewis2020retrieval} provides external knowledge-based inference by retrieving similar patient cases and incorporating them into prompts. 
This approach is particularly appealing for clinical applications, since it enables reasoning without modifying model weights.
Previous studies in content-based image retrieval (CBIR) have shown that case-based reasoning can support dermatological diagnosis by referencing visually similar lesions~\cite{tschandl2019diagnostic,allegretti2021supporting}.
Building on this idea, our work extends case-based reasoning to a multimodal context by integrating retrieved image–text pairs into VLM prompts.
This design supports clinical reasoning by reflecting the way how physicians interpret new cases through analogical comparison with prior patient examples.

More specifically, this study proposes a multimodal diagnostic framework that integrates RAG into a VLM to support more accurate and clinically relevant melanoma classification.
We investigate whether retrieved examples improve diagnostic decisions, particularly in correcting false positives and false negatives.
Through comprehensive experiments, we show that our method consistently outperforms conventional classification models in both accuracy and error correction.
Our main contributions are as follows:
\begin{itemize}[label=\textbullet, itemsep=0pt, topsep=0pt, parsep=0pt, partopsep=0pt]
    \item We propose a retrieval-augmented VLM-based diagnostic framework for mela-noma classification by incorporating image–metadata–label examples into prompts to improve decision accuracy.
    \item We evaluate the effects of different metadata serialization strategies and image encoders on retrieval effectiveness and diagnostic performance.
    \item We show that the proposed method consistently outperforms conventional image-based, text-based, and early-fusion baselines across multiple metrics and architectures, without requiring fine-tuning.
\end{itemize}

\section{Proposed Approach}

This section presents a multimodal diagnostic framework that combines a VLM with RAG to classify melanoma using both dermoscopic images and clinical metadata. In clinical practice, diagnosis often involves comparing a case with prior cases that share similar visual features or clinical attributes. This case-based reasoning improves diagnostic accuracy by using past experience.

\begin{figure}[!t]
    \centering
    \includegraphics[width=\textwidth]{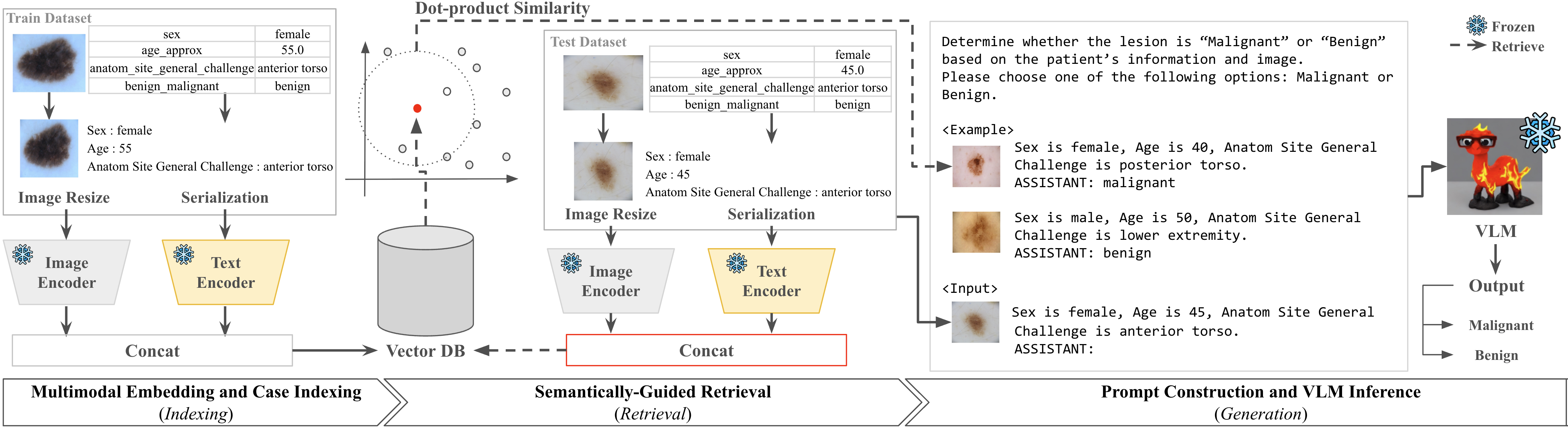}
    \caption{Overview of the proposed retrieval-augmented classification framework incorporating attribute-value pair-based prompting.}
    \label{fig:framework}
\end{figure}

Our framework incorporates a retrieval module that searches a database of dermoscopic images and metadata to find semantically similar cases. These examples serve as clinical references and provide contextual support for the prediction of the model. 
By embedding the retrieved cases into the VLM input prompt, the system emulates the comparative reasoning process used in human diagnosis. 
This design addresses the limitations of general-purpose VLMs and better aligns the model with the specific demands of melanoma classification. An overview of the architecture is shown in Fig.~\ref{fig:framework}.

\vspace{.5em}
\noindent
\textbf{Multimodal Embedding and Case Indexing}
Each training sample includes a dermoscopic image and associated metadata (\textit{e.g.}, age, sex, and lesion location). To process these modalities, we use modality-specific encoders: CNN-based backbones (\textit{e.g.}, ResNeXt-50~\cite{xie2017aggregated}, EfficientNet-V2-M~\cite{tan2019efficientnet}) for images and BERT~\cite{devlin2019bert} for text. Metadata is serialized into natural language using template-based sentence transformations (see below) to improve compatibility with language models.
The resulting image and text embeddings are concatenated into a single multimodal vector and stored in a FAISS~\cite{douze2024faiss} vector database for efficient approximate nearest neighbor search. This setup enables scalable indexing of large dermatological datasets and allows seamless updates as new data is added.

\vspace{.5em}
\noindent
\textbf{Template-Based Sentence Transformations} 
To make structured metadata compatible with VLMs, we convert patient records into natural language using predefined templates. We apply three serialization strategies:
\begin{itemize}[label=\textbullet, itemsep=0pt, topsep=0pt, parsep=0pt, partopsep=0pt]
    \item \textbf{Sentence format}: Expresses each field as a simple sentence (\textit{e.g.}, Age is 45, Sex is female.), matching the VLM’s training style.
    \item \textbf{Attribute-value pair}: Uses compact key-value pairs (\textit{e.g.}, Age: 45, Sex: Female) to reduce prompt length and improve parsing.
    \item \textbf{HTML format}: Encodes tabular structure with tags like <table>, <tr>, <th>, and <td> to retain column semantics.
\end{itemize}

\vspace{.5em}
\noindent
\textbf{Semantically-Guided Retrieval}
At inference time, the query sample (image and metadata) is encoded with the same modality-specific encoders used to index the database. We compute dot-product similarity between the query and database embeddings to retrieve the top-$K$ most similar cases. These examples, selected based on both visual and clinical similarity, provide contextual support for VLM prompting. This retrieval introduces domain-specific knowledge without updating model weights and enables adaptation to the target domain. 
We found that $K = 2$ offers the best balance between contextual relevance and noise.

\vspace{.5em}
\noindent
\textbf{Prompt Construction and VLM Inference}
To adapt general-purpose VLMs for binary melanoma classification, we design structured prompts consists of:
(1) instruction specifying the task;
(2) $K$ retrieved examples as image–metadata–label triplets; and
(3) the query sample with a classification request.

The frozen VLM processes the prompt and generates a textual response indicating the predicted class. This few-shot design mirrors clinical reasoning by analogy and leads to better contextual understanding and more reliable predictions. Unlike early-fusion~\cite{team2024chameleon} or naive multimodal concatenation~\cite{liu2018learn}, our method maintains modality alignment and exploits the ability of the VLM to perform implicit multimodal reasoning.

\begin{table*}[!t]
\caption{Comparison of image(I)-based, metadata(M)-based, early-fusion, zero-shot VLM, and our proposed framework (bold: the highest performance).}
\centering

\resizebox{\textwidth}{!}{

\begin{tabular}{l|cclccccccccccc}

\noalign{\hrule height 1.2pt}
\; & I & M & \;\;\;\;\;\;\; Model & Serialization & Accuracy & Balanced Accuracy & Precision & Sensitivity & F1-score & TN & TP & FN & FP \\
\hline

& \checkmark & - & ResNeXt-50~\cite{xie2017aggregated} & - & 0.7380 & 0.5054 & 0.2022 & 0.1316 & 0.1594 & 6402 & 223 & 1472 & 880 \\
& \checkmark & - & EfficientNet-V2-M~\cite{tan2019efficientnet} & - & 0.6954 & 0.5061 & 0.1985 & 0.2018 & 0.2001 & 5901 & 342 & 1353 & 1381 \\
Single & - & \checkmark & RF~\cite{breiman2001random} & - & 0.8156 & 0.5209 & 0.6667 & 0.0472 & 0.0882 & 7242 & 80 & 1615 & 40 \\
Modality & - & \checkmark & Vicuna 7B v1.5~\cite{zheng2023judging} & HTML & 0.6547 & 0.4873 & 0.1725 & 0.2183 & 0.1927 & 5507 & 370 & 1325 & 1775 \\
& - & \checkmark & Vicuna 7B v1.5 & Attribute-value pair & 0.737 & 0.5263 & 0.2294 & 0.2413 & 0.2352 & 5908 & 409 & 1286 & 1374 \\
& - & \checkmark & Vicuna 7B v1.5 & Sentence & 0.6063 & 0.5152 & 0.2023 & 0.3687 & 0.2613 & 4818 & 625 & 1070 & 2464 \\

\hline

& \checkmark & \checkmark & BERT~\cite{devlin2019bert}+ResNeXt-50 & HTML & 0.8607 & 0.6557 & 0.8366 & 0.3263 & 0.4694 & 7174 & 553 & 1142 & 108 \\
& \checkmark & \checkmark & BERT+ResNeXt-50 & Attribute-value pair & 0.8623 & 0.6568 & 0.8536 & 0.3268 & 0.4277 & 7187 & 554 & 1141 & 95 \\
Early-Fusion & \checkmark & \checkmark & BERT+ResNeXt-50 & Sentence & 0.8622 & 0.6589 & 0.8428 & 0.3322 & 0.4765 & 7177 & 563 & 1132 & 105\\
with RF & \checkmark & \checkmark & BERT+EfficientNet-V2-M & HTML & 0.8501 & 0.6208 & 0.8442 & 0.2525 & 0.3887 & 7203 & 428 & 1267 & 79 \\
& \checkmark & \checkmark & BERT+EfficientNet-V2-M & Attribute-value pair & 0.8501 & 0.6220 & 0.8375 & 0.2555 & 0.3915 & 7198 & 433 & 1262 & 84 \\
& \checkmark & \checkmark & BERT+EfficientNet-V2-M & Sentence & 0.8514 & 0.6232 & \textbf{0.8546} & 0.2566 & 0.3947 & 7208 & 435 & 1260 & 74 \\

\hline

& \checkmark & \checkmark & BERT~\cite{devlin2019bert}+ResNeXt-50 & HTML & 0.6819 & 0.5079 & 0.2000 & 0.2283 & 0.2132 & 5734 & 387 & 1308 & 1548 \\
& \checkmark & \checkmark & BERT+ResNeXt-50 & Attribute-value pair & 0.7040 & 0.5089 & 0.2038 & 0.1953 & 0.1995 & 5989 & 331 & 1364 & 1293 \\
Early-Fusion & \checkmark & \checkmark & BERT+ResNeXt-50 & Sentence & 0.7029 & 0.5009 & 0.1904 & 0.1764 & 0.1832 & 6011 & 299 & 1396 & 1271 \\
with FNN & \checkmark & \checkmark & BERT+EfficientNet-V2-M & HTML & 0.7024 & 0.4967 & 0.1830 & 0.1664 & 0.1743 & 6023 & 282 & 1413 & 1259 \\
& \checkmark & \checkmark & BERT+EfficientNet-V2-M & Attribute-value pair & 0.7084 & 0.5063 & 0.2001 & 0.1817 & 0.1905 & 6051 & 308 & 1387 & 1231 \\
& \checkmark & \checkmark & BERT+EfficientNet-V2-M & Sentence & 0.7108 & 0.5090 & 0.2050 & 0.1847 & 0.1943 & 6068 & 313 & 1382 & 1214 \\

\hline

& \checkmark & \checkmark & LLaVA 7B v1.5 hf~\cite{liu2023visual} & HTML & 0.5845 & 0.6113 & 0.2608 & 0.6543 & 0.3729 & 4138 & 1109 & 586 & 3144 \\
Zero-Shot & \checkmark & \checkmark & LLaVA 7B v1.5 hf & Attribute-value pair & 0.7126 & 0.6128 & 0.3171 & 0.4525 & 0.3729 & 5630 & 767 & 928 & 1652 \\
VLM & \checkmark & \checkmark & LLaVA 7B v1.5 hf & Sentence & 0.5610 & 0.5658 & 0.2320 & 0.5735 & 0.3303 & 4064 & 972 & 723 & 3218 \\

\hline

& \checkmark & \checkmark & BERT+ResNext-50 & HTML & 0.7396 & 0.7202 & 0.3921 & \textbf{0.6891} & 0.4998 & 5471 & 1168 & 527 & 1811 \\
& \checkmark & \checkmark & BERT+ResNext-50 & Attribute-value pair & \textbf{0.8876} & \textbf{0.7970} & 0.7254 & 0.6513 & \textbf{0.6864} & 6864 & 1104 & 591 & 418 \\
& \checkmark & \checkmark & BERT+ResNext-50 & Sentence & 0.8810 & 0.7891 & 0.7027 & 0.6413 & 0.6706 & 6822 & 1087 & 608 & 460 \\
Ours& \checkmark & \checkmark & BERT+EfficientNet-V2-M & HTML & 0.7123 & 0.6746 & 0.3505 & 0.6142 & 0.4463 & 5353 & 1041 & 654 & 1929 \\
($K=2$) & \checkmark & \checkmark & BERT+EfficientNet-V2-M & Attribute-value pair & 0.8491 & 0.7345 & 0.6114 & 0.5504 & 0.5793 & 6689 & 933 & 762 & 593 \\
& \checkmark & \checkmark & BERT+EfficientNet-V2-M & Sentence & 0.8459 & 0.7294 & 0.6022 & 0.4322 & 0.5706 & 6675 & 919 & 776 & 607 \\

\noalign{\hrule height 1.2pt}
\end{tabular} }
\label{table:result}
\end{table*}

\begin{table}[t]
\caption{Comparison of baseline models and the proposed approach (Ours, $K=2$), showing the number of corrected errors (FN/FP corrected as TP/TN) and the corresponding recovery rate (\%). Recovery is defined as the proportion of corrected errors relative to the total baseline errors (I: Image, M: Metadata).}
\centering
\footnotesize
\vspace{.5em}

\subfloat[Performance Comparison Across Different Experimental Settings.]{
\resizebox{\textwidth}{!}{
  \begin{tabular}{l|cclcccc}
  \noalign{\hrule height 1.2pt}
    \toprule
    & I & M & \;\;\;\;\;\; Model & Serialization & & Ours ($K = 2$) & Recovery (\%) \\
    \midrule
     & - & \checkmark & RF~\cite{breiman2001random} & - & FP & 34 & 85.00\\
    Single & - & \checkmark & RF & - & FN & 1035 & 64.09\\
    Modality & - & \checkmark & Vicuna 7B v1.5~\cite{zheng2023judging} & Attribute-value pair & FP & 1258 & 70.87\\
    & - & \checkmark & Vicuna 7B v1.5 & Attribute-value pair & FN & 827 & 64.31\\
    \hline
    Early-Fusion & \checkmark & \checkmark & BERT~\cite{devlin2019bert}+ResNeXt-50~\cite{xie2017aggregated} & Attribute-value pair & FP & 71 & 74.74\\
    with RF & \checkmark & \checkmark & BERT+ResNeXt-50 & Attribute-value pair & FN & 604 & 52.94\\
    \hline
    Zero-Shot & \checkmark & \checkmark & LLaVA 7B v1.5 hf~\cite{liu2023visual} & Attribute-value pair & FP & 1507 & 91.22\\
    VLM & \checkmark & \checkmark & LLaVA 7B v1.5 hf & Attribute-value pair & FN & 571 & 61.53\\
    \bottomrule
    \noalign{\hrule height 1.2pt}
  \end{tabular}
  \label{tab:compare_baseline_recovery}
}}
\hspace{0.01\textwidth} 
\par\vspace{-2.5mm} 

\subfloat[Performance Comparison Across Different Serialization Methods.]{
\resizebox{\textwidth}{!}{
  \begin{tabular}{l|cclcccc}
  \noalign{\hrule height 1.2pt}
    \toprule
    & I & M & \;\;\;\;\;\; Model  & Serialization &  & Attribute-value pair & Recovery (\%) \\
    \midrule
    Ours & \checkmark & \checkmark & BERT~\cite{devlin2019bert}+ResNeXt-50~\cite{xie2017aggregated} & HTML & FP & 1513 & 83.55\\
    ($K = 2$) & \checkmark & \checkmark & BERT+ResNeXt-50 & HTML & FN & 116 & 22.01\\
    \hline
    Ours & \checkmark & \checkmark & BERT+ResNeXt-50 & Sentence & FP & 113 & 24.57\\
    ($K = 2$) & \checkmark & \checkmark & BERT+ResNeXt-50 & Sentence & FN & 43 & 7.07\\
    \bottomrule
    \noalign{\hrule height 1.2pt}
  \end{tabular}
  \label{tab:compare_serialization}
}}
\hspace{0.05\textwidth} 
\par\vspace{-2.5mm} 
\subfloat[Performance Comparison Across Different Image Encoder.]{
\resizebox{\textwidth}{!}{
  \begin{tabular}{l|cclcccc}
  \noalign{\hrule height 1.2pt}
    \toprule
     & I & M & \;\;\;\;\;\; Model  & Serialization & & BERT~\cite{devlin2019bert}+ResNeXt-50~\cite{xie2017aggregated} & Recovery (\%) \\
    \midrule
    Ours & \checkmark & \checkmark & BERT+EfficientNet-V2-M~\cite{tan2019efficientnet} & Attribute-value pair & FP & 487 & 82.12\\
    ($K = 2$) & \checkmark & \checkmark & BERT+EfficientNet-V2-M & Attribute-value pair & FN & 325 & 42.65\\
    \bottomrule
    \noalign{\hrule height 1.2pt}
  \end{tabular}
  \label{tab:tabcompare_image_encoder}
}}
\label{table:all_models}
\end{table}

\section{Experiment}
\noindent
\textbf{Dataset}
We use the SIIM-ISIC 2019 Challenge dataset~\cite{tschandl2018ham10000,codella2018skin,combalia2019bcn20000}, which includes 29,923 dermoscopic images with clinical metadata. Among them, 5,608 cases are histopathologically confirmed melanomas. We treat this as a binary classification task: malignant vs. benign.
Each sample includes an image and metadata (\textit{age}, \textit{sex}, and \textit{anatomical site}). Images are provided in JPEG format and resized to 224×224 RGB. We apply two-stage stratified sampling to preserve class balance: 70\% of the data is used for training and 30\% for testing. The training set is further split 80:20 for validation during hyperparameter tuning.

\vspace{.5em}
\noindent
\textbf{Experimental Setup}
We evaluate performance under five settings: image-based, text-based, multimodal early-fusion, zero-shot VLM, and our retrieval-augmented VLM framework.
For image-based models, we fine-tune ResNeXt-50~\cite{xie2017aggregated} and EfficientNet-V2-M~\cite{tan2019efficientnet}, both initialized with ImageNet weights. Training uses the Adam optimizer with binary cross-entropy loss, and hyperparameters are tuned via Optuna~\cite{akiba2019optuna}.
For text-based classification, we use Random Forest (RF)~\cite{breiman2001random} and 4-bit quantized Vicuna 7B v1.5~\cite{zheng2023judging}, implemented with Scikit-learn~\cite{pedregosa2011scikit} and Hugging Face Transformers~\cite{wolf2020transformers}.
In the early-fusion baseline, we extract image features from the final CNN layer and use the \textit{[CLS]} token from the 11th layer of BERT, following the approach in~\cite{devlin2019bert}.
The two representations are concatenated and classified using either an RF or a ReLU-activated feedforward neural network (FNN). Hyperparameters are tuned via Grid Search~\cite{pedregosa2011scikit} or Optuna~\cite{akiba2019optuna}. For VLM-based experiments, we use the 4-bit quantized version of LLaVA v1.5~\cite{liu2023visual}. In the RAG configuration, we construct a FAISS~\cite{douze2024faiss} index containing 16,756 image–text pairs from the training set. For each query, the top two nearest neighbors ($K=2$) are retrieved and inserted into the model prompt. We empirically evaluate different values of $K$ ($K = 1, 2, 3, 4$) and find that $K=2$ offers the best trade-off between contextual relevance and noise.

\begin{figure}[!t]
  \centering
  \par\vspace{-3mm} 
  \subfloat[Misclassified case by all baselines (LLM, early-fusion, zero-shot VLM) correctly classified by our method ($K=2$).]{
    \includegraphics[width=0.99\textwidth]{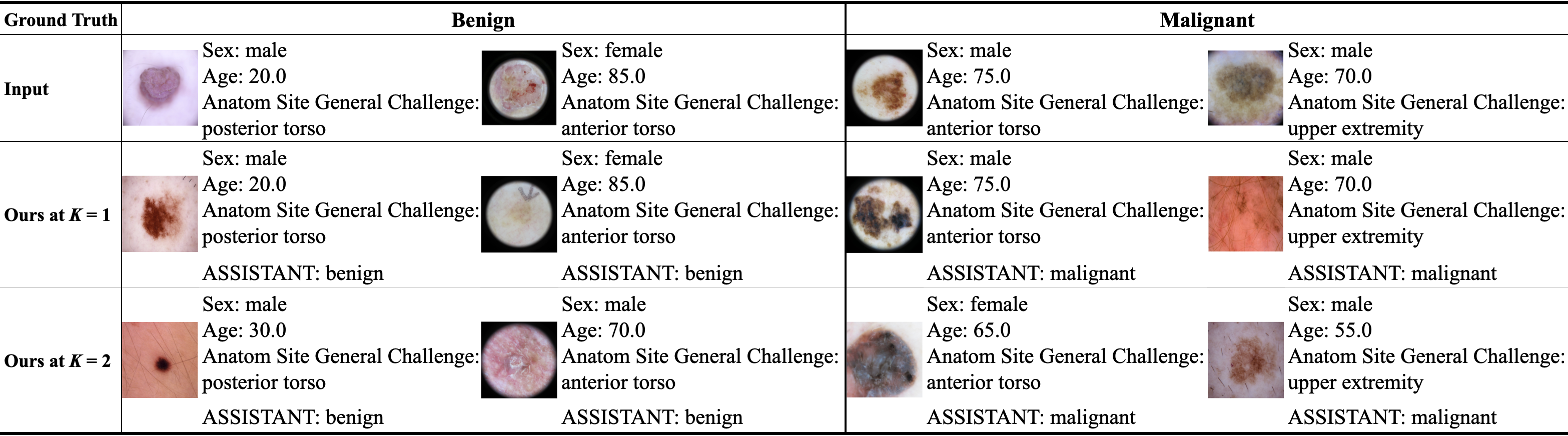}
    \label{fig:compare_baseline_recovery_fig}
  }
  \par\vspace{-3mm} 
  \subfloat[Comparison of HTML and attribute–value formats within RAG framework ($K=2$), showing better results with attribute–value input.]{
    \includegraphics[width=0.99\textwidth]{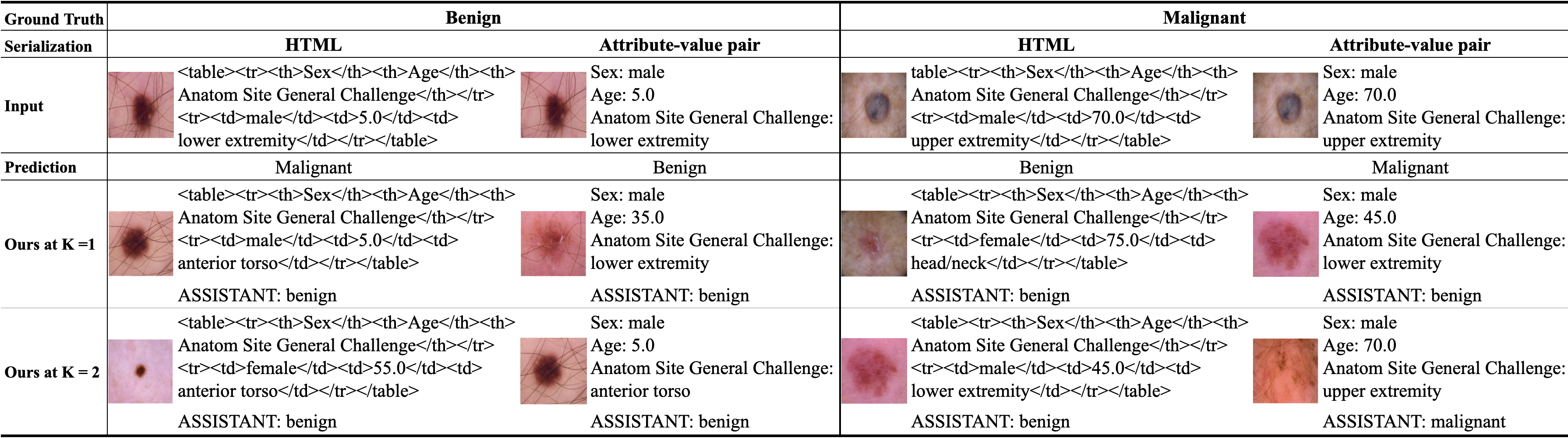}
    \label{fig:compare_serialization_html}
  }
  \par\vspace{-3mm} 
  \subfloat[Comparison of sentence and attribute–value formats, showing improved classification with attribute–value input.]{
    \includegraphics[width=0.99\textwidth]{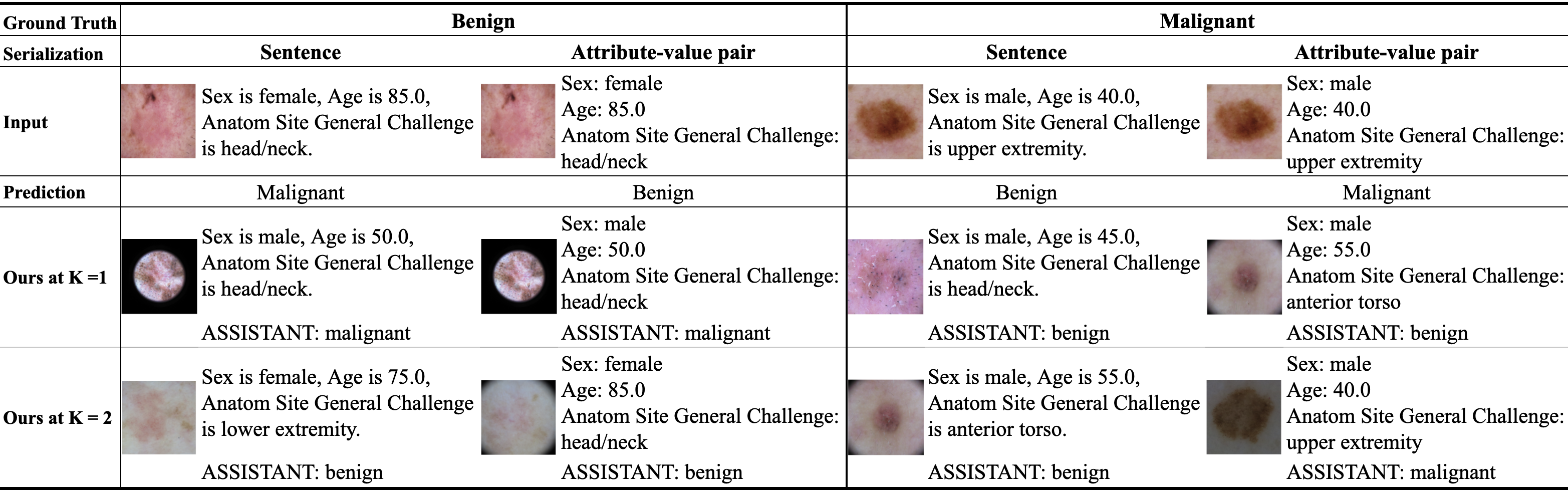}
    \label{fig:compare_serialization_sentence}
  }
  \par\vspace{-3mm} 
  \subfloat[Impact of image encoder choice (EfficientNet-V2-M vs. ResNeXt-50) using attribute–value format in RAG framework.]{
    \includegraphics[width=0.99\textwidth]{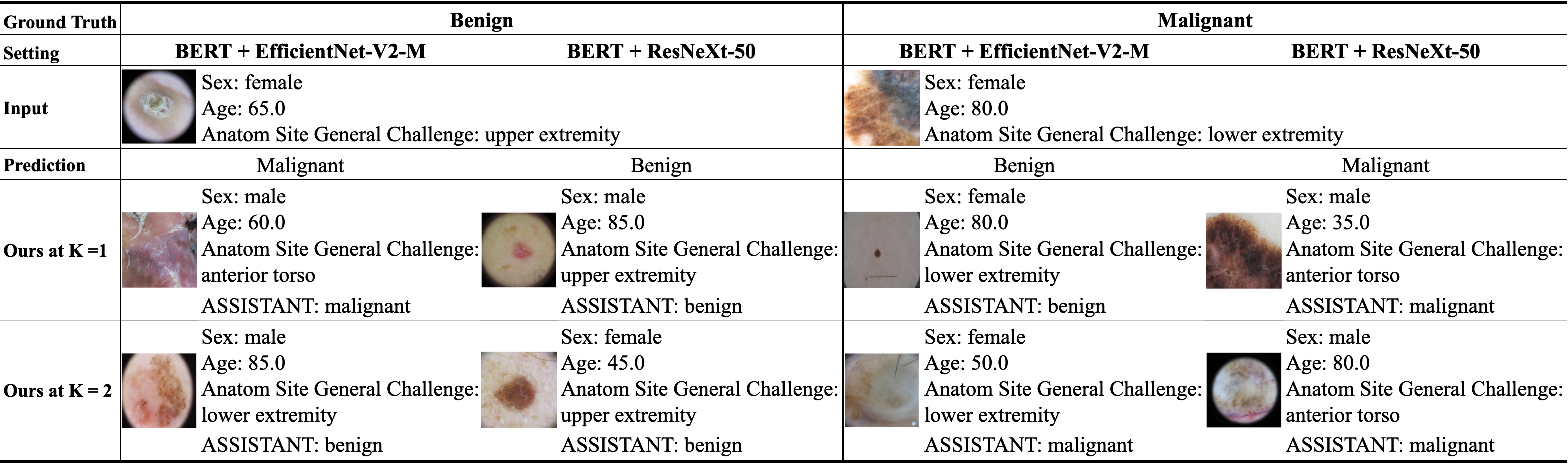}
    \label{fig:compare_encoder_fig}
  }
  \caption{Error cases corrected by our framework with retrieved cases.}
\end{figure}

\subsection{Quantitative Evaluation}
Table~\ref{table:result} summarizes the classification results. We report accuracy, balanced accuracy, precision, sensitivity, F1-score, and confusion matrix components. 
Given the class imbalance and the clinical importance of minimizing both false negative (FN) and false positive (FP), we adopt F1-score as the primary metric.

\vspace{.5em}
\noindent
\textbf{Single-Modality Models}
Models using only dermoscopic images or clinical metadata show limited diagnostic performance. Among image-based methods, EfficientNet-V2-M achieves the best results, though its performance is affected by visual noise due to the absence of preprocessing. In text-based models, Vicuna 7B v1.5 outperforms RF, likely benefiting from general-domain pretraining. 
The RF model struggles to capture patterns effectively due to the limited number of metadata features. 
These results suggest that single-modality approaches are inadequate for accurate melanoma diagnosis and highlight the importance of multimodal integration.

\vspace{.5em}
\noindent
\textbf{Multimodal Fusion and Zero-Shot VLM}
Zero-shot VLM outperforms early-fusion with FNN due to its use of attention-based mechanisms that capture cross-modal interactions. 
In contrast, FNN relies on simple feature concatenation, which limits its ability to represent semantic relationships.
Interestingly, early-fusion with RF achieves better results than zero-shot VLM, indicating that pretrained VLMs do not fully capture clinical signals. 
These observations point to the need for strategies that incorporate domain knowledge and task-specific examples to improve VLM-based classification.

\vspace{.5em}
\noindent
\textbf{Proposed RAG Framework}
Our RAG-based VLM framework achieves the highest performance across all settings. Using BERT, ResNeXt-50, and attribute-value pair serialization, it reaches an F1-score of 0.6864, improving by 0.2099 over the best early-fusion model and by 0.3135 over zero-shot VLM. 
Sensitivity increases to 0.6513, more than doubling that of early-fusion, while precision reaches 0.7254. These results demonstrate that retrieval-augmented prompting provides consistent gains in both accuracy and clinical error correction, while maintaining a strong balance between precision and recall.

\subsection{Qualitative Evaluation}
The quantitative results show that the RAG-based VLM framework outperforms image-based, text-based, early-fusion, and zero-shot VLM models. 
To better understand this performance, we qualitatively examine how retrieved examples contribute to correcting FP and FN that baseline methods fail to resolve.

\vspace{.5em}
\noindent
\textbf{Error Analysis of Baseline Predictions}
Each baseline model uses the best-performing architecture for its modality. Our framework ($K=2$) applies BERT with ResNeXt-50 and attribute–value pair serialization.  
Table~\ref{tab:compare_baseline_recovery} presents recovery rates, defined as the proportion of FP and FN errors that our method correctly reclassifies.  
Zero-shot VLM achieves recovery rates of 91.22\% for FP and 61.53\% for FN.  
In contrast, early-fusion with RF shows substantially lower recovery, likely due to limited capacity to model semantic interactions across modalities.
Fig.~\ref{fig:compare_baseline_recovery_fig} shows representative cases. In the left column, baseline models misclassify benign lesions as malignant. In the right, malignant lesions are predicted as benign.  
Our method retrieves clinically similar examples based on \textit{sex}, \textit{age}, and \textit{anatomical site}, and inserts them into the prompt.  
For instance, two retrieved benign cases from the same \textit{anatomical site} (posterior torso) help correct a prior false positive.  
This context influences the model’s decision and leads to more accurate predictions.

Overall, these results suggest that retrieval-based prompting addresses key limitations of conventional models and supports more reliable clinical reasoning.

\vspace{.5em}
\noindent
\textbf{Effect of Input Serialization Format}  
To evaluate the impact of input serialization, we compare three formats: HTML, attribute–value pair, and sentence, using the same configuration (BERT + ResNeXt-50, $K=2$).  
Table~\ref{tab:compare_serialization} shows recovery rates where FP and FN errors under HTML and sentence formats are corrected to true positive (TP) and true negative (TN) by switching to attribute–value format.

For HTML input, 83.55\% of FP and 22.01\% of FN errors are corrected. In contrast, conversion from sentence format yields 24.57\% for FP and 7.07\% for FN.  
These results suggest that attribute–value input encodes clinical variables more explicitly. Although HTML preserves the same content, its tag-based structure may obscure important features during embedding.  
While sentence format aligns with VLM training data, its classification performance remains lower than attribute–value format.
Fig.~\ref{fig:compare_serialization_html} and~\ref{fig:compare_serialization_sentence} show representative examples.  
In Fig.~\ref{fig:compare_serialization_html}, the HTML-based model yields a false positive for a benign lesion. The attribute–value format retrieves benign cases with matching \textit{sex} and \textit{age} and enables correct classification.  
For the malignant example, the HTML input produces false negatives, whereas attribute–value input retrieves similar malignant lesions and supports accurate prediction.
Fig.~\ref{fig:compare_serialization_sentence} presents a false positive under sentence format that is corrected by attribute–value input, which retrieves benign cases with matching \textit{sex} and \textit{anatomical site}.  
Its more explicit structure strengthens contextual alignment and improves classification.

In summary, these results show that structured input formats, especially attribute–value pairs, better support retrieval-based reasoning by clarifying the semantic role of each variable.

\vspace{.5em}
\noindent
\textbf{Effect of Image Encoder Configuration}
To assess the influence of image encoder on diagnostic performance, we compare ResNeXt-50 and EfficientNet-V2-M, using the same text encoder (BERT), input format (attribute-value pair), and retrieval setting ($K=2$).
Table~\ref{tab:tabcompare_image_encoder} shows recovery rates where predictions by the EfficientNet-V2-M model are corrected by the ResNeXt-50 model. The recovery rate for FPs reaches 82.12\%, and for FNs 42.65\%, indicating that ResNeXt-50 is more effective at correcting errors.
Fig.~\ref{fig:compare_encoder_fig} presents examples where the two models produce different outcomes for the same inputs. In the benign case, EfficientNet-V2-M misclassifies the lesion as malignant. In contrast, ResNeXt-50 retrieves benign cases with similar \textit{anatomical site} and \textit{sex}, which results in correct classification. These retrieved cases show strong alignment with the query in both spatial and demographic attributes. EfficientNet-V2-M, by comparison, retrieves cases with lower consistency, which may reduce contextual reliability.
A similar difference appears in the malignant case. EfficientNet-V2-M predicts FN, whereas ResNeXt-50 retrieves two malignant cases with similar color and lesion spread, resulting in correct classification. These examples provide clearer visual evidence that supports the decision of the model.

Altogether, the results suggest that ResNeXt-50 captures visual features relevant to melanoma classification more effectively and provides stronger contextual alignment in retrieval-based inference.

\section{Conclusion}

We presented a retrieval-augmented diagnostic framework that integrates VLMs with case-based prompting for melanoma classification.  
The framework improves diagnostic performance without fine-tuning by retrieving semantically similar examples and inserting them into the input prompt.  
Quantitative and qualitative results show improved contextual reasoning and fewer classification errors.  
Comparisons across serialization formats and encoders confirm the benefit of structured input and ResNeXt-50 visual features.
These findings support retrieval-augmented prompting as a robust and generalizable strategy for clinical decision support using pretrained multimodal models.  
Although the framework shows promising results, its reliance on a single VLM may limit generalizability across diverse diagnostic tasks.  
Future work may expand this approach to multi-class skin lesion classification and other domains that require multimodal reasoning and greater model flexibility.

\begin{credits}
\subsubsection{\ackname} This research was supported (1) by the MSIT (Ministry of Science and ICT), Korea, under the Global Research Support Program in the Digital Field (RS-2024-00431394), supervised by the IITP (Institute for Information \& Communications Technology Planning \& Evaluation); (2) by the MSIT and NIPA (National IT Industry Promotion Agency), Korea, under the Development of AI Precision Medical Solution project (Dr. Answer 2.0, S0252-21-1001); and (3) under the High Performance Computing Support project.

\end{credits}
\bibliographystyle{splncs04}
\bibliography{Retrieval-Augmented_VLMs_for_Multimodal_Melanoma_Diagnosis}
\end{document}